\documentclass{article}

\usepackage{times}

\usepackage{graphicx} 
\usepackage{subfigure} 

\usepackage{natbib}

\usepackage{algorithm}
\usepackage{algorithmic}

\usepackage{hyperref}

\renewcommand{\algorithmicrequire}{\textbf{Assume:}}

\usepackage[accepted]{icml2014}

\usepackage{amsmath}
\usepackage{amssymb}
\usepackage{xspace}
\usepackage{listings}

\definecolor{darkgreen}{rgb}{0,.4,0}
\definecolor{orange}{rgb}{.8,.4,0}
\newcommand{\Normal}{\mathcal{N}}

\newcommand{\Discrete}{\text{Discrete}}
\newcommand{\E}{\mathbb{E}}

\newcommand{\x}{\+x_{1:N}}
\newcommand{\y}{y_{1:N}}

\newcommand{\observe}{\texttt{\footnotesize observe}\xspace}
\newcommand{\predict}{\texttt{\footnotesize predict}\xspace}

\newcommand{\fork}{\texttt{\footnotesize fork}\xspace}
\newcommand{\program}[1]{\texttt{\footnotesize #1}}
\newcommand{\POSIX}{POSIX\xspace}

\begin{document} 

\twocolumn[
\icmltitle{A Compilation Target for Probabilistic Programming Languages}

\icmlauthor{Brooks Paige}{brooks@robots.ox.ac.uk}
\icmlauthor{Frank Wood}{fwood@robots.ox.ac.uk}
\icmladdress{University of Oxford, Department of Engineering Science, Oxford, UK}

\icmlkeywords{probabilistic programming, compilation, fork}

\vskip 0.3in
]


\begin{abstract} 
Forward inference techniques such as sequential Monte Carlo and particle Markov chain Monte Carlo for probabilistic programming can be implemented 
in any programming language 
by creative use of standardized operating system functionality including processes, forking, mutexes, and shared memory.   
Exploiting this we have defined, developed, and tested a probabilistic programming language intermediate representation language we call probabilistic C,
 which itself can be compiled to machine code by standard compilers and linked to operating system libraries yielding an efficient, scalable, portable probabilistic programming compilation target.  This opens up a new hardware and systems research path for optimizing probabilistic programming systems.
\end{abstract}

\section{Introduction}
\label{sec:introduction}

Compilation is source to source transformation. 
We use the phrase intermediate representation to refer to a target language for a compiler.  
This paper introduces a C-language library that makes possible a C-language intermediate representation for probabilistic programming languages that can itself be compiled to executable machine code. 
We call this intermediate language {\em probabilistic C}.  
Probabilistic C can be compiled normally and uses only macros and \POSIX operating system libraries \cite{posix} to implement general-purpose, scalable, parallel probabilistic programming inference.  
Note that in this paper we do not show how to compile any existing probabilistic programming languages (i.e.~IBAL~\citep{pfeffer2001ibal}, BLOG \citep{milch20071}, Church \citep{goodman2008church}, Figaro \citep{pfeffer2009figaro}, Venture \citep{venture}, or Anglican \citep{wood2014anglican}) to this intermediate representation; instead we leave this to future work noting there is a wealth of readily available resources on language to language compilation that could be leveraged to do this.  
We instead characterize the performance of the intermediate representation itself by writing programs directly in probabilistic C and then testing them on computer architectures and programs that illustrate the capacities and trade-offs of both the forward inference strategies probabilistic C employs and the operating system functionality on which it depends.  

Probabilistic C programs compile to machine executable meta-programs that perform inference over the original program via forward methods such as sequential Monte Carlo \cite{doucet2001sequential} and particle MCMC variants \cite{andrieu2010particle}.  Such inference methods can be implemented in a sufficiently general way so as to support inference over the space of probabilistic program execution traces using just \POSIX operating system primitives.

\subsection{Related work}

The characterization of probabilistic programming inference we consider here is the process of sampling from the a posteriori distribution of  execution traces arising from stochastic programs constrained to reflect observed data.  This is the view taken by the Church \cite{goodman2008church},
 Venture \cite{venture},
 and Anglican \cite{wood2014anglican},
 programming languages among others.
In such languages, models for observed data can be described purely in terms of a forward generative process.

Markov chain Monte Carlo (MCMC) is used by these systems to sample from the posterior distribution of program execution traces.
Single-site Metropolis Hastings (MH) \cite{goodman2008church} and particle MCMC (PMCMC) \cite{wood2014anglican} are two such approaches.
In the latter it was noted that a \program{fork}-like operation is a fundamental requirement of forward inference methods for probabilistic programming, where \program{fork} is the standard \POSIX operating system primitive \citep{posixfork}.   \citet{kiselyov09} also noted that delimited continuations, a user-level generalization of \program{fork} could be used for inference, albeit in a restricted family of models. 
%
%

\section{Probabilistic Programming}
\label{sec:programs}

\begin{figure}
\begin{lstlisting}
#include "probabilistic.h"

int main(int argc, char **argv) {

    double var = 2;
    double mu = normal_rng(1, 5);

    observe(normal_lnp(9, mu, var)); 
    observe(normal_lnp(8, mu, var)); 
    
    predict("mu,%f\n", mu);

    return 0;
}
\end{lstlisting}
\caption{
This program performs posterior inference over the unknown mean \program{mu} of a Gaussian,
conditioned on two data points.
The \predict directive
formats output using standard \program{printf} semantics.
}
\label{prog:gaussian}
\end{figure}

\begin{figure}[tb]
\begin{lstlisting}
#include "probabilistic.h"
#define K 3
#define N 11

/* Markov transition matrix */
static double T[K][K] = { { 0.1,  0.5,  0.4 }, 
                          { 0.2,  0.2,  0.6 }, 
                          { 0.15, 0.15, 0.7 } };

/* Observed data */
static double data[N] = { NAN, .9, .8,  .7,   0, -.025,
                               -5, -2, -.1,   0,  0.13 };

/* Prior distribution on initial state */
static double initial_state[K] = { 1.0/3, 1.0/3, 1.0/3 };

/* Per-state mean of Gaussian emission distribution */
static double state_mean[K] = { -1, 1, 0 };

/* Generative program for a HMM */
int main(int argc, char **argv) {

    int states[N];
    for (int n=0; n<N; n++) {
        states[n] = (n==0) ? discrete_rng(initial_state, K) 
                           : discrete_rng(T[states[n-1]], K);
        if (n > 0) {
            observe(normal_lnp(data[n], state_mean[states[n]], 1));
        }
        predict("state[%d],%d\n", n, states[n]);
    }
    
    return 0;
}
\end{lstlisting}
\caption{A hidden Markov model (HMM) with 3 underlying states and Gaussian emissions, observed at 10 discrete time-points.
We observe 10 data points and predict the marginal distribution over the latent \program{state} at each time point.
}
\label{prog:hmm}
\end{figure}

Any program that makes a random choice over the course of its execution implicitly defines a prior distribution over its random variables;
running the program can be interpreted as drawing a sample from the prior.
Inference in probabilistic programs involves conditioning on observed data, and characterizing the posterior distribution of the random variables given data.
We introduce probabilistic programming capabilities into C by providing a library with two primary functions:
\observe which conditions the program execution trace given the log-likelihood of a data point, and
\predict which marks expressions for which we want posterior samples.
%
Any random number generator and sampling library can be used for making random choices in the program, any numeric log likelihood value can be passed to an \observe,
and any C expression which can be printed can be reported using \predict.  
The library includes a single macro which renames \texttt{\footnotesize main} and wraps it in another function that runs the original in an inner loop in the forward inference algorithms to be described.

Although C is a comparatively low-level language, it can nonetheless represent many well-known generative models concisely and transparently.
Figure~\ref{prog:gaussian} shows a simple probabilistic C program for estimating the posterior distribution for the mean of a Gaussian, conditioned on two observed data points $y_1, y_2$, corresponding to the model
\begin{align}
\mu &\sim \Normal(1, 5), 
&
y_1, y_2 &\stackrel{iid}{\sim} \Normal(\mu, 2).
\end{align}
We \observe the data $y_1,y_2$ and \predict the posterior distribution of $\mu$.
The functions \program{normal\_rng} and \program{normal\_lnp} in Figure~\ref{prog:gaussian} return (respectively) a normally-distributed random variate 
and the log probability density of a particular value, with mean and variance parameters \program{mu} and \program{var}.
The \observe statement requires only the log-probability of the data points $8$ and $9$ conditioned on the current program state; no other information about the likelihood function or the generative process.
In this program we predict the posterior distribution of a single value \program{mu}.

A hidden Markov model example is shown in Figure~\ref{prog:hmm}, in which $N = 10$ observed data points $y_{1:N}$ are drawn from an underlying Markov chain
with $K$ latent states, each with Gaussian emission distributions with mean $\mu_k$, and a (known) $K\times K$ state transition matrix $T$, such that
\begin{align}
z_0 &\sim \Discrete([1/K, \dots, 1/K]) \\
z_n|z_{n-1} &\sim \Discrete(T_{z_{n-1}}) \\
y_n|z_n &\sim \Normal(\mu_{z_n}, \sigma^2).
\end{align}
Bayesian nonparametric models can also be represented concisely;
in Figure~\ref{prog:crp} we show a generative program for an infinite mixture of Gaussians.
We use a Chinese restaurant process (CRP) to sequentially sample non-negative integer partition assignments $z_n$
for each data point $y_1, \dots, y_N$.
For each partition, mean and variance parameters $\mu_{z_n}, \sigma_{z_n}^2$ are drawn from a normal-gamma prior;
the data points $y_n$ themselves are drawn from a normal distribution with these parameters, defining a full generative model
\begin{align}
z_n &\sim \mathrm{CRP}(\alpha, z_1, \dots, z_{n-1}) \\
{1}/{\sigma_{z_n}^2} &\sim \mathrm{Gamma}(1, 1) \\
\mu_{z_n}|\sigma_{z_n}^2 &\sim \Normal(0, \sigma_{z_n}^2) \\
y_n|z_n, \mu_{z_n}, \sigma_{z_n}^2  &\sim \Normal(\mu_{z_n}, \sigma_{z_n}^2).
\end{align}
This program also demonstrates the additional library function \program{memoize}, which can be used to implement stochastic memoization as described in \cite{goodman2008church}.

\begin{figure}[t]
\begin{lstlisting}
#include "probabilistic.h"
#define N 10

// Observed data
static double data[N] = { 1.0,  1.1,   1.2,
                         -1.0, -1.5,  -2.0, 
                        0.001, 0.01, 0.005, 0.0 };

// Struct holding mean and variance parameters for each cluster
typedef struct theta {
    double mu;
    double var;
} theta;

// Draws a sample of theta from a normal-gamma prior
theta draw_theta() {
    double variance = 1.0 / gamma_rng(1, 1);
    return (theta) { normal_rng(0, variance), variance };
}

// Get the class id for a given observation index
static polya_urn_state urn;
void get_class(int *index, int *class_id) {
    *class_id = polya_urn_draw(&urn);
}

int main(int argc, char **argv) {
    double alpha = 1.0;
    polya_urn_new(&urn, alpha);

    mem_func mem_get_class; 
    memoize(&mem_get_class, get_class, sizeof(int), sizeof(int));

    theta params[N];
    bool known_params[N] = { false };

    int class;
    for (int n=0; n<N; n++) {
        mem_invoke(&mem_get_class, &n, &class);
        if (!known_params[class]) {
            params[class] = draw_theta();
            known_params[class] = true;
        }
        observe(normal_lnp(data[n], params[class].mu, 
                                    params[class].var));
    }

    // Predict number of classes
    predict("num_classes,%2d\n", urn.len_buckets);

    // Release memory; exit
    polya_urn_free(&urn);
    return 0;
}
\end{lstlisting}
\caption{A infinite mixture of Gaussians on the real line.
Class assignment variables
for each of the 10 data points
 are drawn following a Blackwell-MacQueen urn scheme to sequentially sample from a Dirichlet process.
}
\label{prog:crp}
\end{figure}

\subsection{Operating system primitives}

Inference proceeds by drawing posterior samples from the space of program execution traces.
We define an execution trace as the sequence of memory states 
(the entire virtual memory address space)
that arises during the sequential step execution of machine instructions.

The algorithms we propose for inference in probabilistic programs map directly onto standard computer operating system constructs, exposed in \POSIX-compliant operating systems including Linux, BSD, and Mac OS X.
The cornerstone of our approach is \POSIX \fork \cite{posixfork}.
When a process forks, it clones itself, creating a new process with an identical copy of the execution state of the original process, and identical source code;
both processes then continue with normal program execution completely independently from the point where \fork was called.
While copying program execution state may na{\"i}vely sound like a costly operation, this actually can be rather efficient:
when \fork is called, a lazy copy-on-write procedure is used to avoid deep copying the entire program memory.
Instead, initially only the pagetable is copied to the new process; when an existing variable is modified in the new program copy, then and only then are memory contents duplicated.
The overall cost of forking a program is proportional to the fraction of memory which is rewritten by the child process \cite{Smith88effectsof}.

Using \fork we can branch a single program execution state and explore many possible downstream execution paths.
Each of these paths runs as its own process, and will run in parallel with other processes.
In general, multiple processes run in their own memory address space, and do not communicate or share state. 
We handle inter-process communication via a small shared memory segment; the details of what global data must be stored are provided later.

Synchronization between processes is handled via mutual exclusion locks ({\em mutex} objects).
Mutexes become particularly useful for us when used in conjunction with a synchronized counter to create a {\em barrier},
a high-level blocking construct which prevents any process proceeding in execution state beyond the barrier until some fixed number of processes
have arrived.

\section{Inference}
\label{sec:inference}

\subsection{Probability of a program execution trace}

To notate the probability of a program execution trace, we enumerate all $N$ \observe statements, and the associated observed data points $y_1, \dots, y_N$.
During a single run of the program, some total number $N'$ random choices $\+x'_1, \dots, \+x'_{N'}$ are made.
While $N'$ may vary between individual executions of the program, 
we require that the number of \observe directive calls $N$ is constant. 

The observations $y_n$ can appear at any point in the program source code 
and define a partition of the random choices $\+x'_{1:N'}$ into $N$ subsequences $\+x_{1:N}$,
where each $\+x_n$ contains all random choices made up to observing $y_n$ but excluding any random choices prior to observation $y_{n-1}$.
We can then define the probability of any single program execution trace 
\begin{align}
p(\y, \x) &= \prod_{n=1}^N g(y_n|\+x_{1:n}) f(\+x_n|\+x_{1:n-1})
\end{align}
In this manner, any model with a generative process that can be written in C code with stochastic choices can be represented in this sequential form in the space of program execution traces.

Each \observe statement takes as its argument $\ln g(y_n|\+x_{1:n})$.
Each quantity of interest in a \predict statement corresponds to some function $h(\cdot)$ of all random choices $\x$ made during the execution of the program.
Given a set of $S$ posterior samples $\{\x^{(s)}\}$, we can approximate the posterior distribution of the \predict value as
\begin{align}
h(\x) \approx \frac{1}{S} \sum_{s=1}^S h(\x^{(s)}).
\end{align}

\subsection{Sequential Monte Carlo}

Forward simulation-based algorithms are a natural fit for probabilistic programs: run the program and report executions that match the data.
Sequential Monte Carlo (SMC, sequential importance resampling) forms the basic building block of other, more complex particle-based methods,
and 
can itself be used as a simple approach to probabilistic programming inference.
SMC approximates a target density $p(\x|\y)$ 
as a weighted set of $L$ realized trajectories $\x^{\ell}$ such that
\begin{align}
p(\x|\y) &\approx \sum_{\ell = 1}^L w^\ell_N \delta_{\x^\ell}(\x).
\end{align}
For most probabilistic programs of interest, 
it will be intractable to sample from $p(\x|\y)$ directly. 
Instead, noting that (for $n > 1$) we have the recursive identity 
\begin{align}
p(&\+x_{1:n}|y_{1:n}) \\ \nonumber
&= p(\+x_{1:n-1}|y_{1:n-1}) g(y_n|\+x_{1:n}) f(\+x_n|\+x_{1:n-1}),
\end{align}
we sample from $p(\x|\y)$ by iteratively sampling from each
$p(\+x_{1:n}|y_{1:n})$, in turn, from $1$ through $N$.
At each $n$, we construct an importance sampling distribution by proposing from some
distribution $q(\+x_n|\+x_{1:n-1}, y_{1:n})$;
in probabilistic programs we find it convenient to propose directly from the executions of the program, i.e.~each
sequence of random variates $\+x_n$ is jointly sampled from the program execution state dynamics 
\begin{align}
\+x_n^\ell \sim f(\+x_n|\+x_{1:n-1}^{a_{n-1}^\ell})
\end{align}
where $a_{n-1}^\ell$ is an ``ancestor index,'' the particle index $1, \dots, L$ of the parent (at time $n-1$) of $\+x_n^\ell$.
The unnormalized particle importance weights at each observation $y_n$ are simply the \observe data likelihood
\begin{align}
\tilde w_n^\ell = g( y_{1:n}, \+x_{1:n}^\ell)
\end{align}
which can be normalized as
\begin{align}
w_n^\ell &= \frac{\tilde w_n^\ell}{\sum_{\ell=1}^L \tilde w_n^\ell}.
\end{align}
After each step $n$, we now have a weighted set of execution traces which approximate $p(\+x_{1:n}|y_{1:n})$.
As the program continues, traces which do not correspond well with the data will have weights which become negligibly small, leading in the worst case to all weight concentrated on a single execution trace.
To counteract this deficiency, we resample from our current set of $L$ execution traces after each observation $y_n$, according to their weights $w_n^\ell$.
This is achieved by sampling a count $O^\ell_n$ for the number of ``offspring'' of a given execution trace $\ell$ to be included at time $n+1$;
any sampling scheme must ensure $\E[O^\ell_n] = w_n^\ell$.
Sampling offspring counts $O^\ell_n$ is equivalent to sampling ancestor indices $a_n^\ell$.
Program execution traces with no offspring are killed; program execution traces with more than one offspring are forked multiple times.
After resampling, all weights $w_n^\ell = 1$.

\begin{algorithm}[tb]
\caption{Parallel SMC program execution}
\label{algo:smc}
\begin{algorithmic}
\REQUIRE 
$N$ observations, $L$ particles
\STATE launch $L$ copies of the program \hfill (parallel)
\FOR{$n=1\dots N$}
	\STATE wait until all $L$ reach \observe $y_n$ \hfill (barrier)
	\STATE update unnormalized weights $\tilde w^{1:L}_n$ \hfill (serial)
	\IF{$ESS < \tau$}
		\STATE sample number of offspring $O_n^{1:L}$ \hfill (serial)
		\STATE set weight $\tilde w^{1:L}_n = 1$ \hfill (serial)
		\FOR{$\ell=1\dots L$}
			\STATE fork or exit \hfill (parallel)
		\ENDFOR
	\ELSE
		\STATE set all number of offspring $O_n^{\ell} = 1$ \hfill (serial)
	\ENDIF
	\STATE continue program execution \hfill (parallel)
\ENDFOR
\STATE wait until $L$ program traces terminate \hfill (barrier)
\STATE \predict from $L$ samples from $\hat p(\x^{1:L}|\y)$ \hfill (serial)
\end{algorithmic}
\end{algorithm}

We only resample if the effective sample size 
\begin{align}
ESS 
\approx 
\frac{1}{\sum_\ell (w_n^{\ell})^2}
\end{align}
is less than some threshold value $\tau$; we choose $\tau = L/2$.

In probabilistic C, each \observe statement forms a barrier: parallel execution cannot continue until all particles have arrived at the \observe and have reported their current unnormalized weight.
As execution traces arrive at the \observe barrier, they take the number of particles which have already reached the current \observe as a (temporary) unique identifier.
Program execution is then blocked as the effective sample size is computed and the number of offspring are sampled.
The number of offspring are stored in a shared memory block;
when the number of offspring are computed, each particle uses the identifier assigned when reaching the \observe barrier to retrieve (asynchronously) from shared memory the number of children to \fork.
Particles with no offspring wait for any child processes to complete execution, and terminate; particles with only one offspring do not \fork any children but continue execution as normal.

The SMC algorithm is outlined in Algorithm~\ref{algo:smc}, with annotations for which steps are executed in parallel, serially, or form a barrier.
After a single SMC sweep is complete, we sample values for each $\predict$, and then (if desired) repeat the process, running a new independent particle filter, to draw an additional batch of samples.


\subsection{Particle Metropolis-Hastings}

Particle Markov chain Monte Carlo, introduced in \citet{andrieu2010particle}, uses sequential Monte Carlo to generate high-dimensional proposal distributions for MCMC.
The most simple formulation is the particle independent Metropolis-Hastings algorithm.
After running a single particle filter sweep, we compute an estimate of the marginal likelihood,
\begin{align}
\hat Z \equiv p(\y) \approx \prod_{n=1}^N \left [ \frac{1}{L} \sum_{\ell = 1}^L w_n^\ell \right ].
\end{align}
We then run another iteration of sequential Monte Carlo which we use as a MH proposal; we estimate the marginal likelihood $\hat Z'$ of the new proposed particle set, and then with probability $\min(1, \hat Z' / \hat Z)$ we accept the new particle set and output a new set of \predict samples, otherwise outputting the same \predict samples as in the previous iteration.

The inner loop of Algorithm~\ref{algo:pimh} is otherwise substantially similar to SMC.

\begin{algorithm}[tb]
\caption{Parallel PIMH program execution}
\label{algo:pimh}
\begin{algorithmic}
\REQUIRE $M$ iterations, $N$ observations, $L$ particles
\FOR{$m=1 \dots M$}
\STATE launch $L$ copies of the program \hfill (parallel)
\FOR{$n=1\dots N$}
	\STATE wait until all $L$ reach an \observe \hfill (barrier)
	\STATE update unnormalized weights $\tilde w_{1:L}$ \hfill (serial)
	\IF{$ESS < \tau$}
		\STATE update proposal evidence estimate $\hat Z'$  \hfill (serial)
		\STATE sample number of offspring $O_n^{1:L}$ \hfill (serial)
		\STATE set weight $\tilde w^{1:L}_n = 1$ \hfill (serial)
		\FOR{$\ell=1\dots L$}
			\STATE fork or exit \hfill (parallel)
		\ENDFOR
	\ELSE
		\STATE set all number of offspring $O_n^{\ell} = 1$ \hfill (serial)
	\ENDIF
	\STATE continue program execution \hfill (parallel)
\ENDFOR
\STATE wait until $L$ program traces terminate \hfill (barrier)
\STATE accept or reject new particle set \hfill (serial)
\STATE \predict from $L$ samples from $\hat p(\x^{1:L}|\y)$ \hfill (serial)
\STATE store current particle set ${\+x}$ and evidence $\hat Z$ \hfill (serial)
\STATE continue to next iteration \hfill (parallel)
\ENDFOR
\end{algorithmic}
\end{algorithm}

\subsection{Particle Gibbs}

\begin{algorithm}[tb]
\caption{Parallel Particle Gibbs program execution}
\label{algo:pmcmc}
\begin{algorithmic}
\REQUIRE $M$ iterations, $N$ observations, $L$ particles
\FOR{$m=1 \dots M$}
\STATE $L' \leftarrow L$ if $m = 1$, otherwise $L-1$
\STATE launch $S'$ copies of the program \hfill (parallel)
\FOR{$n=1\dots N$}
	\STATE wait until all $L'$ reach an \observe \hfill (barrier)
	\STATE compute weights for all particles \hfill (serial)
	\IF{$m > 1$}
		\STATE signal num offspring to retained trace \hfill (serial)
	\ENDIF
	\FOR{$\ell=1\dots L'$}
		\STATE spawn retain / branch process [Algo.~\ref{algo:retain-and-branch}] \hfill (parallel)
	\ENDFOR
   	\STATE wait until $L$ particles finish branching \hfill (barrier)
	\STATE continue program execution \hfill (parallel)
\ENDFOR
\STATE wait until $L$ program traces terminate \hfill (barrier)
\STATE \predict from $L$ samples from $\hat p(\x^{1:L}|\y)$ \hfill (serial)
\STATE select and signal particle to retain \hfill (serial)
\STATE wait until $N$ processes are ready to branch \hfill (barrier)
\STATE continue to next iteration \hfill (parallel)
\ENDFOR
\end{algorithmic}
\end{algorithm}

\begin{algorithm}[tb]
\caption{Retain and Branch inner loop}
\label{algo:retain-and-branch}
\begin{algorithmic}
\REQUIRE input initial $C > 0$ children to spawn
\STATE \program{is\_retained} $\leftarrow$ \program{false}
\WHILE{\program{true}}
\IF{$ C = 0$ and not \program{is\_retained}}
	\STATE discard this execution trace, exit
\ELSE[$C \ge 0$]
	\STATE spawn $C$ new children
\ENDIF
\STATE wait for signal which resets \program{is\_retained}
\IF{\program{is\_retained}}
	\STATE wait for signal which resets $C$
\ELSE
	\STATE discard this execution trace, exit
\ENDIF
\ENDWHILE
\end{algorithmic}
\end{algorithm}

Particle Gibbs is a particle MCMC technique which also has SMC at its core, with better theoretical statistical convergence properties than PIMH but additional computational overhead.
We initialize particle Gibbs by running a single sequential Monte Carlo sweep, and then alternate between
(1)
 sampling a single execution trace $\hat{\+x}_{1:M}$ from the set of $L$ weighted particles, and
(2) 
 running a ``conditional'' SMC sweep, in which we generate $L-1$ new particles in addition to the retained $\hat{\+x}_{1:M}$.

The implementation based on operating system primitives is described in algorithms~\ref{algo:pmcmc}~and~\ref{algo:retain-and-branch}.
The challenge here is that we must ``retain'' an execution trace, which we can later revisit to resume and branch arbitrarily many times.
This is achieved by spawning off a ``control'' process at every observation point, which from then on manages the future of that particular execution state.

As before, processes arrive at an \observe barrier, and when all particles have reached the observe we compute weights, and sample offspring counts $O_n^\ell$.
Particles with $O_n^\ell = 0$ terminate, but new child processes are no longer spawned right away.
Instead, all remaining particles \fork a new process whose execution path immediately diverges from the main codebase and enters the retain and branch loop in Algorithm~\ref{algo:retain-and-branch}.
This new process takes responsibility for actually spawning the $O_n^\ell$ new children.
The spawned child processes (and the original process which arrived at the \observe barrier) wait (albeit briefly) at a new barrier marking the {\em end} of \observe $n$, not continuing execution until all new child processes have been launched.

Program execution continues to the next \observe, during which the retain / branch process waits until a full particle set reaches the end of the program.
Once final weights $\tilde w_N^{1:L}$ are computed, we sample (according to weight) from the final particle set to select a single particle to retain during the next SMC iteration.
When the particle is selected, a signal is broadcast to all retain / branch loops indicating which process ids correspond to the retained particle;
all except the retained trace exit.

The retain / branch loop now goes into waiting again (this time for a branch signal), and we begin SMC from the top of the program.
As we arrive at each observe $n$, we only sample $L-1$ new offspring to consider: we guarantee that at least one offspring is spawned from the retained particle at $n$ (namely, the retained execution state at $n+1$).
However, depending on the weights, often sampling offspring will cause us to want more than a single child from the retained particle.
So, we signal to the retained particle execution state at time $n$ the number of children to spawn; the retain / branch loop returns to its entry point and resumes waiting, to see if the previously retained execution state will be retained yet again.

Note that in particle Gibbs, we must resample (select offspring and reset weights $w_n^\ell = 1$) after every observation
in order to be able to properly align the retained particle on the next iteration through the program.

%
%
%
%
%

\section{Experiments}
\label{sec:experiments}

We now turn to benchmarking probabilistic C against existing probabilistic programming engines, and evaluate the relative strengths of the inference algorithms in Section~\ref{sec:inference}.  
We find that compilation improves performance by approximately 100 times over interpreted versions of the same inference algorithm.  
We also find evidence that suggests that optimizing operating systems to support probabilistic programming usage could yield significant performance improvements as well.

The programs and models we use in our experiments are chosen to be sufficiently simplistic that we can compute the exact posterior distribution of interest analytically, allowing us to evaluate correctness of inference.
Given the true posterior distribution $p$, we measure sampler performance by the KL-divergence $KL(\hat p||p)$, where $\hat p$ is our Monte Carlo estimate.
The first benchmark program we consider is a hidden Markov model (HMM) very similar to that of Figure~\ref{prog:hmm}, where we predict the marginal distributions of each latent state.
The HMM used in our experiments here is larger; it has the same model structure, but with $K=10$ states and 50 sequential observations,
and each state $k = 1, \dots, 10$ has a Gaussian emission distribution with $\mu_k = k-1$ and $\sigma^2 = 4$.
The second benchmark is the CRP mixture of Gaussians program in Figure~\ref{prog:crp}, where we predict the total number of distinct classes.

\subsection{Comparative performance of inference engines}

\begin{figure*}[t]
\centering
\includegraphics[width=\columnwidth]{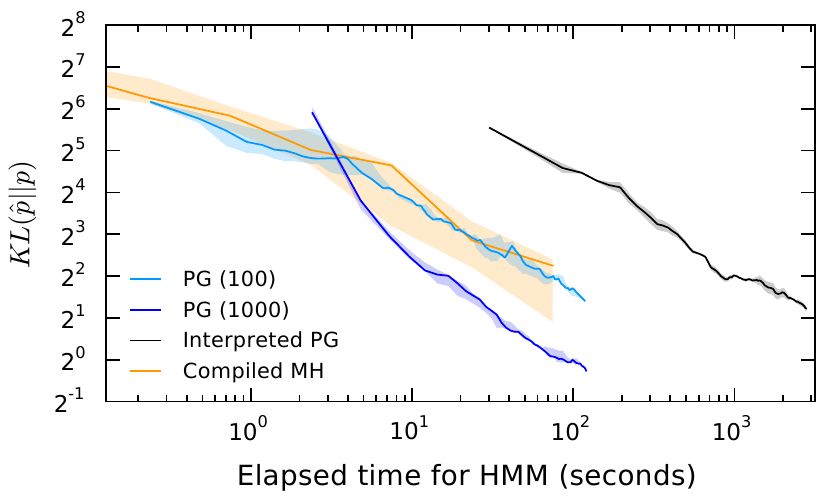}
\includegraphics[width=\columnwidth]{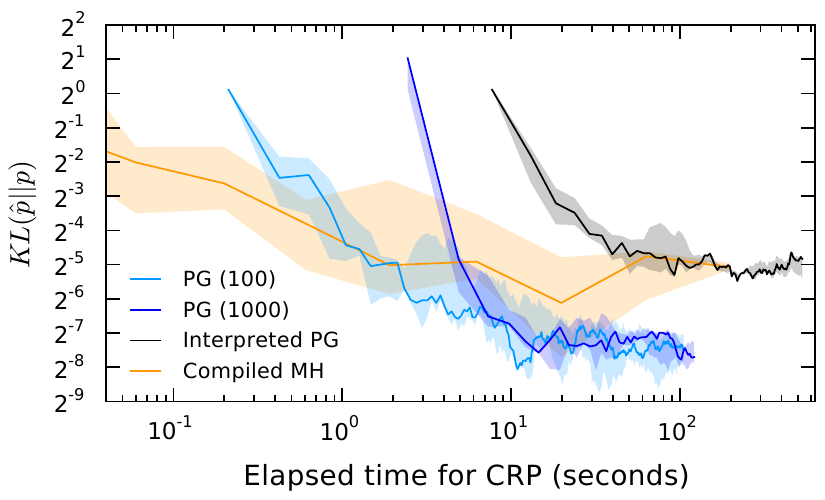}
\caption{Performance plots for (left) HMM and (right) CRP models, run in an 8 core computing environment on Amazon EC2.
In both models we see that compiling particle Gibbs, compared to running within an interpreter, leads to a large (approx.~$100\times$) constant-factor speed increase.
The Metropolis-Hastings sampler converges at a similar rate as particle Gibbs with 100 particles in the HMM, but appears to mix slower asymptotically for the CRP.
In both models, increasing to 1000 particles in particle Gibbs yields somewhat faster convergence  at the expense of a longer waiting time until the first sample set arrives.
}
\label{fig:crp-and-hmm}
\end{figure*}

\begin{figure*}[tbh]
\centering
\includegraphics[width=\columnwidth]{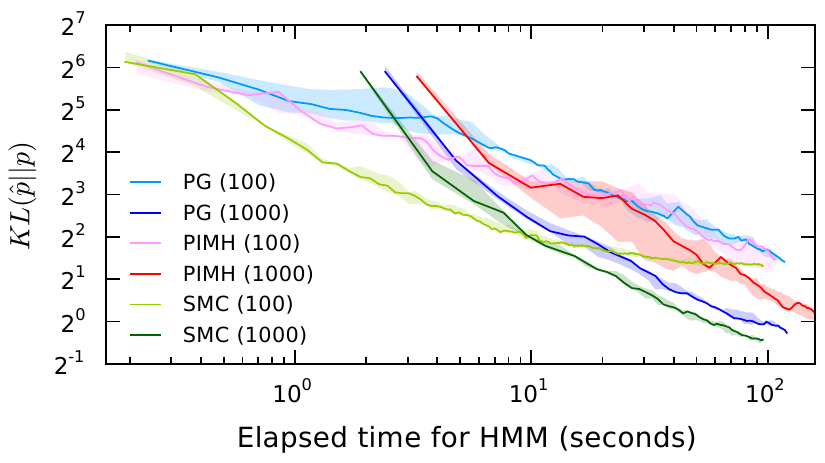}
\includegraphics[width=\columnwidth]{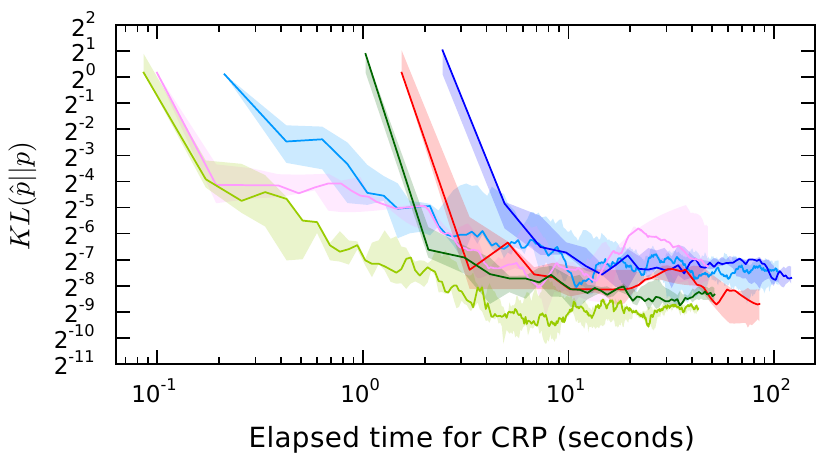}
\caption{Comparison of SMC, PIMH, and PG for 100 and 1000 particles in (left) the HMM, and (right) the CRP.
The relative computational and statistical efficiency of the PIMH and PG algorithms varies across models and number of particles. 
The SMC algorithm is quick to draw initial samples, but has no guarantee of convergence to the target density; we see this in its apparently poor asymptotic behavior in the HMM with 100 particles.
Colors are the same across plots.
}
\label{fig:inference-types}
\vspace{-.1em}
\end{figure*}

We begin by benchmarking against two existing probabilistic programming engines: {\em Anglican}, as described in \citet{wood2014anglican}, which also implements particle Gibbs but is an interpreted language based on Scheme, implemented in Clojure, and running on the JVM; and {\em probabilistic-js}\footnote{https://github.com/dritchie/probabilistic-js}, a compiled system implementing the inference approach in \citet{wingate2011lightweight}, which runs Metropolis-Hastings over each individual random choice in the program execution trace.
The interpreted particle Gibbs engine is multithreaded, and we run it with 100 particles and 8 simultaneous threads;
the Metropolis-Hastings engine only runs on a single core.
In Figure~\ref{fig:crp-and-hmm} we compare inference performance in both of these existing engines to our particle Gibbs backend, running with 100 and 1000 particles, in an 8 core cloud computing environment on Amazon EC2, running on Intel Xeon E5-2680 v2 processors.
Our compiled probabilistic C implementation of particle Gibbs runs over $100$ times faster that the existing interpreted engine,
generating good sample sets in on the order of tens of seconds.

The probabilistic C inference engine implements particle Gibbs, SMC, and PIMH sampling, which we compare in Figure~\ref{fig:inference-types} using both 100 and 1000 particles.
SMC is run indefinitely by simply repeatedly drawing independent sets of particles; in contrast to the PMCMC algorithms, this is known to be a biased estimator even as the number of iterations goes to infinity \cite{whiteley2013}, and is not recommended as a general-purpose approach.

Figures~\ref{fig:crp-and-hmm} and \ref{fig:inference-types} plot wall clock time against KL-divergence.
We use all generated samples as our empirical posterior distribution in order to produce as fair a comparison as possible.
In all engines, results are reasonably stable across runs; the shaded band covers the 25th to 75th percentiles over multiple runs, with the median marked as a dark line.
A sampler drawing from the target density will show approximately linear decrease in KL-divergence on these log-log plots; a steeper slope correspond to greater statistical efficiency.
The methods based on sequential Monte Carlo do not provide any estimate of the posterior distribution until completing a single initial particle filter sweep;
for large numbers of particles this may be a non-trivial amount of time.
In contrast, the MH sampler begins drawing samples effectively immediately, although it may take a large number of individual steps before converging to the correct stationary distribution; individual Metropolis-Hastings samples are likely to be more autocorrelated, but producing each one is faster.


%
%
%
%

\vspace{-0.07cm}
\subsection{Performance characteristics across multiple cores}

\begin{figure}[tb]
\centering
\includegraphics[width=\columnwidth]{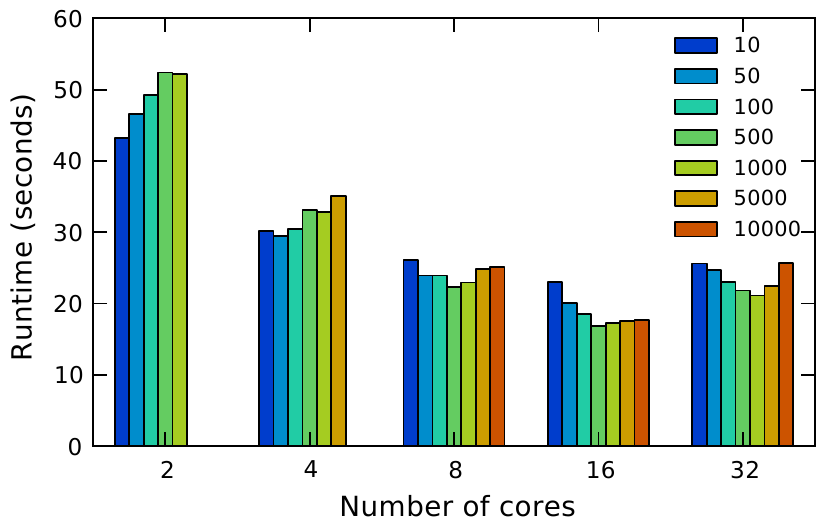}
\vspace{-.55cm}
\caption{Effect of system architecture on runtime performance.  Here we run the HMM code on EC2 instances with identical processors (horizontal axis) with varying number of particles (individual bars) and report runtime to produce 10,000 samples.  Despite adding more cores, after 16 cores performance begins to degrade. Similarly, adding more particles eventually degrades performance for any fixed number of cores. In combination these suggest the availability of operating system optimizations that could improve overall performance.
}
\label{fig:add-cores}
\end{figure}

As the probabilistic C inference engine offloads much of the computation to underlying operating system calls, 
we characterize the limitations of the OS implementation by comparing time to completion as we vary the number of cores.
Tests for the hidden Markov model across core count (all on EC2, all with identical Intel Xeon E5-2680 v2 processors) are shown in Figure~\ref{fig:add-cores}.

%
%

%
%

%
%

\section{Discussion}
\label{sec:discussion}

Probabilistic C is a method for performing inference in probabilistic programs.   
Methodologically it derives from the forward methods for performing inference in statistical models based on sequential Monte Carlo and particle Markov chain Monte Carlo.   
We have shown that it is possible to efficiently and scalably implement this particular kind of inference strategy using existing, standard compilers and \POSIX compliant operating system primitives.  
 
What most distinguishes Probabilistic C from prior art is that it is highly compatible with modern computer architectures, all the way from operating systems to central processing units (in particular their virtual memory operations), and, further, that it delineates a future research program for scaling the performance of probabilistic programming systems to large scale problems by investigating systems optimizations of existing computer architectures.  
Note that this is distinct but compatible with approaches to optimizing probabilistic programming systems by compilation optimizations, stochastic hardware, and dependency tracking with efficient updating of local execution trace subgraphs.  
It may, in the future, be possible to delineate model complexity and hardware architecture regimes in which each approach is optimal; we assert that, for now, it is unclear what those regimes are or will be.  
This paper is but one step towards such a delineation.

Several interesting research questions remain: 
(1) Is it more sensible to write custom memory management and use threads than fork and processes as we have done?   The main contribution of this paper is to establish a probabilistic programming system implementation against a standardized, portable abstraction layer.  It might be possible to eke out greater performance by capitalizing on the fact modern architectures are optimised for parallel threads more so than parallel processes; however, exploiting this would entail implementing memory management de facto equivalent in action to fork which may lead to lower portability. 
 (2) Would shifting architectures to small page sizes help?  There is a bias towards large page size support in modern computer architectures.  It may be that the system use characteristics of probabilistic programming systems might provide a counterargument to that bias or inspire the creation of tuned end-to-end systems.  Forking itself is lightweight until variable assignment which usually require manipulations of entire page tables.  Large pages require large amounts of amortisation in order to absorb the cost of copying upon stochastic variable assignment.  Smaller pages could potentially yield higher efficiencies.  
 (3) What characteristics of process synchronisation can be improved specifically for probabilistic programming systems?  This is both a systems and machine learning question.  From a machine learning perspective we believe it may be possible to construct efficient sequential Monte Carlo algorithms that do not synchronize individual threads at observe barriers and instead synchronize in a queue.  On a systems level it begs questions about what page replacement strategies to consider; perhaps entirely changing the page replacement schedule to reflect rapid process rather than thread multiplexing across cores.

\vspace{0.325cm}
Probabilistic C 
does not disallow programmers from accidentally writing programs 
that are statistically incoherent.  
Many probabilistic programming languages (including Church, Anglican, and Venture) are nearly purely functional and so being disallow program variable value reassignment.
This ensures a well-defined joint distribution on program variables.
Probabilistic C offers no such guarantees.
For this reason we are not, in this paper, making a claim about a new probabilistic programming language per se --- rather we describe probabilistic C as an intermediate representation compilation target that implements a particular style of inference that is natively parallel and possible to optimize by system architecture choice.  
It is possible (as helpfully pointed out by 
Vikash Mansinghka)
that compiler optimization techniques such as checking to see whether or not the program is natively in ``static single assignment'' form \cite{appel1998ssa} can help avoid statistically incoherent programs being allowed (at compilation).  
This we leave as future work.
Further (as helpfully pointed out by 
Noah Goodman),
programming languages constructs such as delimited continuations \cite{felleisen1988theory} can be thought of as user level abstractions of fork, and might provide similar functionality at the user rather than system level in the context of statistically safer languages. 

\pagebreak

\bibliography{refs.bib}
\bibliographystyle{icml2014}

\end{document}